\documentclass[letterpaper, 10 pt, conference]{ieeeconf}  

\IEEEoverridecommandlockouts                              

\title{\LARGE \bf
    SCA3D: Enhancing Cross-modal 3D Retrieval via \\ 3D Shape and Caption Paired Data Augmentation
}

\author{$\text{Junlong Ren}^{\dag}$, $\text{Hao Wu}^{\dag}$, Hui Xiong, Hao Wang*
\thanks{$^{\dag}{\text{Equal contribution}}$; *Corresponding author.}
\thanks{J. Ren, H. Wu, H. Xiong, H. Wang are with AI Thrust, The Hong Kong University of Science and Technology (Guangzhou), Guangzhou,
        China. Email: 
{\tt\small jren686@connect.hkust-gz.edu.cn, hwubx@connect.ust.hk, xionghui@hkust-gz.edu.cn, haowang@hkust-gz.edu.cn}.
        }%
}

\usepackage{cite}
\usepackage{amsmath,amssymb,amsfonts}
\usepackage{algorithmic}
\usepackage{graphicx}
\usepackage{textcomp}
\usepackage{xcolor}
\usepackage{spconf,amsmath,graphicx}
\usepackage{multirow}
\usepackage{tabularray}
\usepackage{subfig}
\usepackage{algorithm}
\usepackage{algorithmic}
\usepackage{amssymb}
\usepackage{booktabs}
\usepackage{colortbl}
\usepackage{pifont}       
\usepackage{bbding}       
\usepackage{hyperref}

\begin{document}

\maketitle
\thispagestyle{empty}
\pagestyle{empty}

\begin{abstract}
The cross-modal 3D retrieval task aims to achieve mutual matching between text descriptions and 3D shapes. 
This has the potential to enhance the interaction between natural language and the 3D environment, especially within the realms of robotics and embodied artificial intelligence (AI) applications.
However, the scarcity and expensiveness of 3D data constrain the performance of existing cross-modal 3D retrieval methods. These methods heavily rely on features derived from the limited number of 3D shapes, resulting in poor generalization ability across diverse scenarios. 
To address this challenge, we introduce SCA3D, a novel 3D shape and caption online data augmentation method for cross-modal 3D retrieval. Our approach uses the LLaVA model to create a component library, captioning each segmented part of every 3D shape within the dataset. Notably, it facilitates the generation of extensive new 3D-text pairs containing new semantic features. 
We employ both inter and intra distances to align various components into a new 3D shape, ensuring that the components do not overlap and are closely fitted. Further, text templates are utilized to process the captions of each component and generate new text descriptions. Besides, we use unimodal encoders to extract embeddings for 3D shapes and texts based on the enriched dataset. We then calculate fine-grained cross-modal similarity using Earth Mover's Distance (EMD) and enhance cross-modal matching with contrastive learning, enabling bidirectional retrieval between texts and 3D shapes.
Extensive experiments show our SCA3D outperforms previous works on the Text2Shape dataset, raising the Shape-to-Text RR@1 score from 20.03 to 27.22 and the Text-to-Shape RR@1 score from 13.12 to 16.67.
Codes can be found in \href{https://github.com/3DAgentWorld/SCA3D}{https://github.com/3DAgentWorld/SCA3D}.

\end{abstract}

\section{INTRODUCTION}
In robotics perception \cite{huang2023visual,li2024manipllm,huang2024copa, hu2024look}, retrieval plays a crucial role as the information gathered supports the following robot control and behavior. With the increasing complexity of robots including drones and underwater vehicles, the perception of the 3D world has become more critical. Researches in this area increasingly focus on the perception of 3D environments. Moreover, as studies on 3D vision tasks \cite{zhang2024multigo,wang2024embodiedscan,bakrcot3dref,man2024lexicon3d, song2025gvkf,cheng2025graphguided,yu2025rgb} advance, the quality of 3D perception improves significantly. This enhancement enables robots to acquire and interpret more comprehensive information from the 3D world. Consequently, advancements in 3D retrieval are crucial in helping robots to perceive and understand the real world more effectively.

Cross-modal 3D retrieval provides robots with a method to interact with the 3D world through natural language, highlighting its importance in robotics. Previous works in cross-modal 3D retrieval \cite{chen2019text2shape, han2019y2seq2seq, ruan2024tricolo, tang2021parts2words, wu2024com3d} primarily focus on the combination of 3D features from geometry in 3D shapes and text embeddings from textual data. Various matching methods are employed to align these 3D and text features. While these techniques have shown promising results on simulated 3D-text datasets, they encounter challenges when faced with more complex, real-world data.  

In this paper, we introduce a novel online data augmentation method for the cross-modal 3D retrieval task. 
The performance of 3D retrieval tasks is often limited by simplistic synthetic datasets. The absence of real-world 3D-text datasets poses significant challenges for these models in various applications. 
To address this limitation, we utilize LLaVA \cite{liu2024visual} to caption segmented parts within the limited 3D shapes in the 3D-text dataset, thereby constructing an abundant component library of 3D shape parts with rich textual descriptions.
Based on the component library, our proposed online data augmentation method allows the generation of vast 3D-text paired data from a minimal set of real examples. 

We align various components into new 3D shapes by applying both inter-component and intra-component distance adjustments, ensuring the components are closely fitted together without any overlap. Moreover, text templates are used to handle the captions of each component, producing new text descriptions that match the newly created 3D shapes. This capability significantly enhances performance by providing extensive data support crucial for applications in realistic scenarios such as robotic perception where labeled data are scarce.
Besides, we use Earth Mover's Distance (EMD) to compute fine-grained cross-modal similarity for the alignment between 3D shapes and text descriptions. Considering the effects of data augmentation, we also incorporate contrastive learning and adopt InfoNCE loss \cite{oord2018representation} to enhance the effectiveness of cross-modal alignment.

\begin{figure*}[t]
\centering
  \subfloat{\includegraphics[width=1\linewidth]{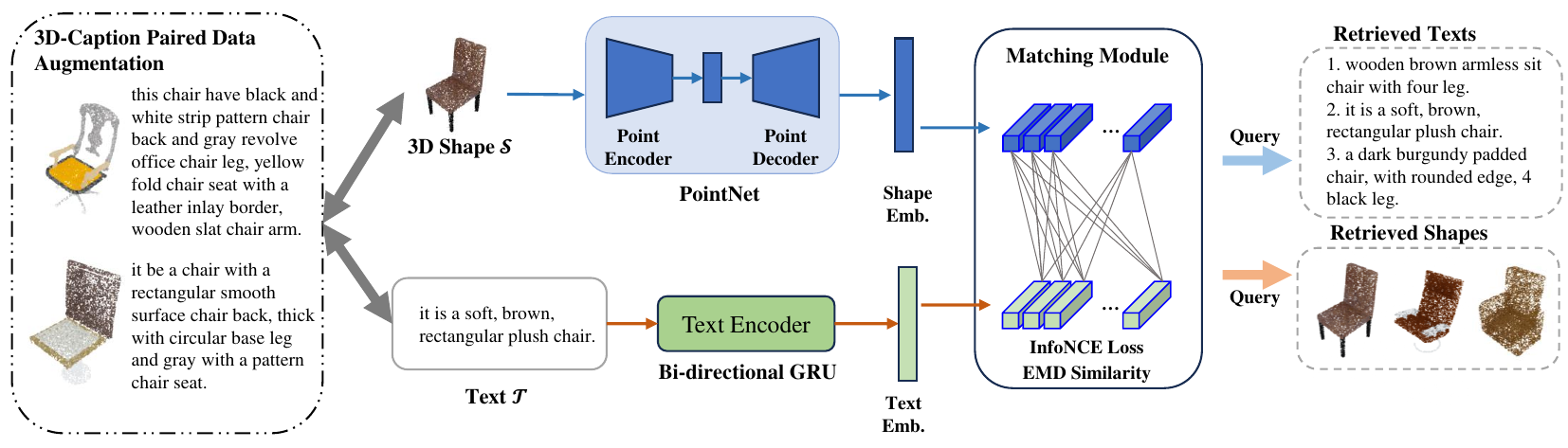}} 
   
\caption{\textbf{The overview of our proposed SCA3D.} It consists of three components: the 3D-caption paired data augmentation module, unimodal encoders, and the matching module. The 3D-caption paired data augmentation module continuously creates extensive 3D-text pairs with diverse geometry and semantics to facilitate cross-modal training. The unimodal encoders comprise a 3D shape encoder and a text encoder, which learn 3D shape and text embeddings from the input data. The matching module computes similarity scores between each 3D-text pair using Earth Mover's Distance (EMD), maximizing the similarity of positive pairs while minimizing the similarity of negative pairs.}
\label{figure::overview}
\vspace{-1.0em}
\end{figure*}

In summary, the main contributions of our paper are:
\begin{itemize}
    \item We introduce a novel online data augmentation method for cross-modal 3D retrieval capable of generating vast 3D-text paired data. This approach alleviates the issue of data scarcity and significantly enhances data diversity.
    \item  We implement cross-modal 3D-text pairing in data augmentation. This allows our method to modify semantics and improve robustness across varied scenarios.
    \item Extensive experiments demonstrate that our SCA3D surpasses existing methods on the Text2Shape dataset by achieving significant improvements. It raises the Shape-to-Text (S2T) RR@1 score from 20.03 to 27.22 and the Text-to-Shape (T2S) RR@1 score from 13.12 to 16.67, showcasing superior performance and robust generalization capabilities.
\end{itemize}

\section{RELATED WORK}

\subsection{2D-Text Matching}
In recent years, 2D-text matching models such as CLIP \cite{radford2021learning}, BLIP \cite{li2022blip}, and Open-VCLIP \cite{weng2023open} have demonstrated impressive performance not only on retrieval tasks but also across numerous downstream tasks. The success is primarily attributed to the availability of large-scale image-text and video-text pretraining datasets like LAION-400M \cite{schuhmann2021laion} and HowTo100M \cite{miech2019howto100m}.
In particular, CLIP \cite{radford2021learning} pre-trained on 400M image-text pairs achieves remarkable zero-shot performance across 27 datasets, including ImageNet \cite{deng2009imagenet}. 

Recent methods \cite{Peng_2024_CVPR, doveh2023teaching} also leverage the generation capabilities of diffusion models \cite{podellsdxl} and large language models (LLMs) \cite{le2022bloom} for data augmentation. However, these methods do not generate data during training due to the high computational cost of diffusion models and LLMs, limiting the diversity of augmented data.
In contrast, we generate part-level captions using a multimodal large language model (MLLM) and then randomly sample multiple parts to compose complex 3D shapes with corresponding captions during the training process, introducing minimal additional computational cost. This approach ensures the diversity of both shape geometry and text semantics, leading to more robust and effective data augmentation.

\subsection{3D-Text Matching}
Text2Shape \cite{chen2019text2shape} introduces a 3D-text dataset by captioning 3D shapes from ShapeNet \cite{chang2015shapenet} and proposes a framework to learn joint embeddings of 3D shapes and natural languages. This framework consists of a 3D-CNN and GRU \cite{chung2014empirical} to encode 3D voxelized shapes and texts, followed by metric learning to achieve alignment between modalities. $\rm Y^{2}$Seq2Seq \cite{han2019y2seq2seq} models both multi-view images and texts in a sequence-to-sequence manner to jointly reconstruct and predict view and word sequences. TriCoLo \cite{ruan2024tricolo} proposes a trimodal training framework to jointly align 3D voxels, multi-view images, and texts.
Parts2Words \cite{tang2021parts2words} employs regional-based matching to compute local similarities and enhance retrieval performance. COM3D \cite{wu2024com3d} further considers cross-view correspondence and augments 3D features using SRT \cite{sajjadi2022scene}. However, these methods primarily focus on extracting more discriminative cross-modal representations, overlooking the scarcity of 3D-text paired data. We try to mitigate this issue by applying data augmentation with an MLLM to extensively create new 3D-text pairs, leading to robust and generalized retrieval capability.

In addition to the aforementioned 3D-text retrieval methods, PointCLIP \cite{zhang2022pointclip} and CLIP2Point \cite{huang2023clip2point} train additional adapters with depth maps to transfer 2D CLIP knowledge to 3D shape classification. Nevertheless, they do not effectively bridge the gap between 2D and 3D visual information including self-occlusion, due to the limited number of multi-view images. We adopt point clouds as 3D shape representations to better model geometric information.

\begin{figure*}[t]
\centering
  \subfloat{\includegraphics[width=1\linewidth]{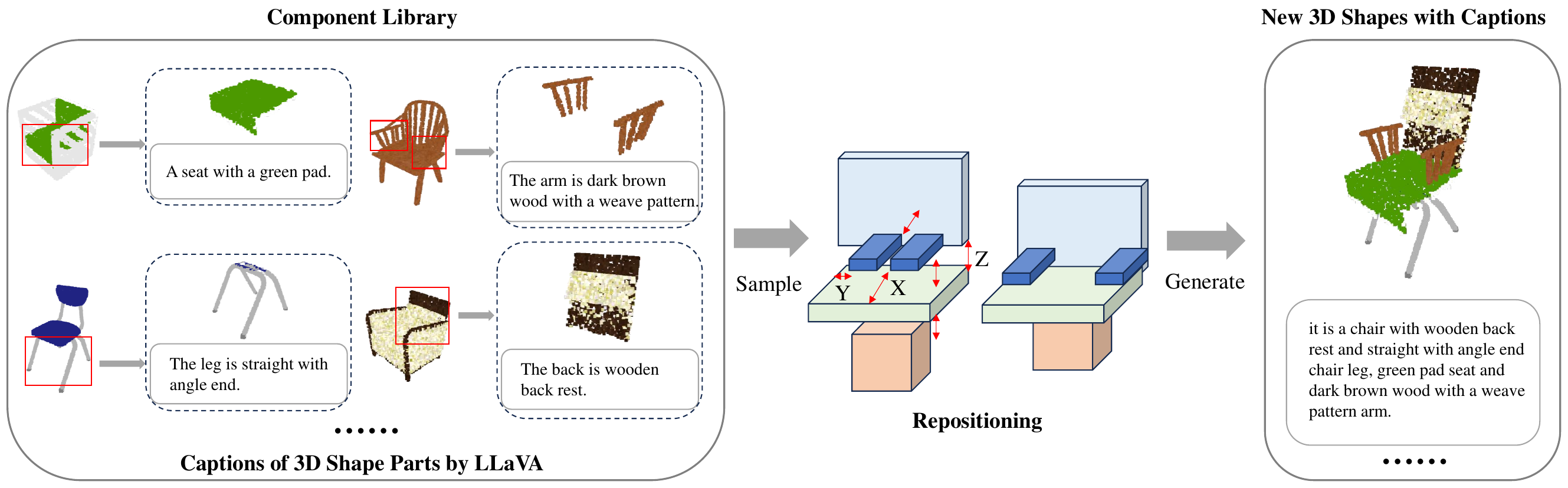}} 
   
\caption{\textbf{The pipeline of 3D-caption paired data augmentation.} The component library is created by captioning 3D shape parts through LLaVA. During training, different components are sampled from this library, and repositioning is applied to generate new 3D shapes with correct geometry and corresponding text captions.}
\label{figure::data augmentation}
\end{figure*}

\section{METHODOLOGY}
\subsection{Overview}
Our cross-modal 3D retrieval framework comprises three components: the data augmentation module, the unimodal encoders, and the matching module. To achieve data-efficient cross-modal 3D retrieval, the data augmentation module samples multiple parts from different 3D shapes to create diverse new shapes with accurate captions. The unimodal encoders include a 3D shape encoder and a text encoder, which encode a 3D shape $S$ and a text caption $T$ into the embedding space of 3D shape and language modalities.
For cross-modal matching between $S$ and $T$, the matching module scores each pair of $S$ and $T$ using Earth Mover's Distance (EMD), and it is optimized by contrastive learning. The overview of our framework is shown in Fig. \ref{figure::overview}. 

\subsection{3D-Caption Paired Data Augmentation}

The scarcity of 3D assets and the high cost of human annotation have constrained the scale of 3D-text datasets. Pretraining on large-scale datasets has been proven effective for 2D-text matching methods \cite{radford2021learning,li2022blip, weng2023open}. Therefore, we aim to enrich current 3D-text datasets using an MLLM as an annotator. Specifically, we obtain part-level captions instead of shape-level captions, as 3D parts can be easily reassembled into integrated shapes. Captioning parts enables the generation of new shapes with corresponding shape-level captions. 
The permutation of parts also facilitates generating large-scale 3D-text pairs with diverse geometry and semantics.
The pipeline of data augmentation is illustrated in Fig. \ref{figure::data augmentation}.

\paragraph{Captioning 3D Shapes in Part-level} To generate new shapes and corresponding text captions, we first caption each part of every 3D shape in the training set. The Text2Shape \cite{chen2019text2shape} dataset in cross-modal 3D retrieval shares the same 3D models with the segmentation dataset PartNet \cite{mo2019partnet}. The annotations in PartNet define how a shape can be semantically segmented (e.g., a table can be segmented into a tabletop and a table base). Given the predefined semantic segmentation labels and shape-level captions, we leverage an LLM to generate captions for each part. 
We adopt an MLLM, i.e. LLaVa \cite{liu2024visual}, instead of a unimodal LLM to utilize the visual information of 3D shapes. Concretely, we render 3D shapes into multi-view images, then prompt the MLLM with these images, shape-level captions, and semantic part types. To enrich vocabulary diversity, the MLLM is also prompted to output captions that cover as many phrases and words as possible. Finally, we obtain a library of components consisting of part-level shapes and captions.

\begin{algorithm}[t]
	\renewcommand{\algorithmicrequire}{\textbf{Input:}}
	\renewcommand{\algorithmicensure}{\textbf{Output:}}
	\caption{3D-Caption Paired Data Generation Process}
	\label{alg1}
	\begin{algorithmic}[1]
        \REQUIRE component library $L$, caption template $Tem$
        \ENSURE generated 3D shape $S$ and text caption $T$

        \STATE Sample $N$ parts and texts $\left. \left\{ \left({p}_{n}, {t}_{n}\right) \right\}_{n = 1}^{N} \right.\in L$
        \STATE Initialize distance matrix $D \in \mathbb{R}^{N \times N}$

        \FOR{$i \leftarrow 1$ \textbf{to} N}
            \STATE  $p_i \leftarrow $ Reposition centroid to the origin
            \STATE Fill in the template $Tem.fill\left(t_i\right)$
        \ENDFOR

        \FOR{$i \leftarrow 1$ \textbf{to} N}
            \FOR{$j \leftarrow 1$ \textbf{to} N}
            \STATE  $d_{ij} \leftarrow $ Compute marginal distances on XYZ axes
            \ENDFOR
        \ENDFOR

        \FOR{$i \leftarrow 1$ \textbf{to} N}
            \STATE  $p_i \leftarrow $ Adjust inter and intra distances with $\left\{d_{ij}\right\}_{j=1}^{N}$
        \ENDFOR

        \STATE $S \leftarrow \{p_1; p_2; \cdots; p_n\}, T \leftarrow Tem$

        \RETURN $S, T$

	\end{algorithmic}  
\end{algorithm}

\paragraph{Generating 3D-Caption Paired Data} To generate new shapes, we randomly select a shape category (e.g., a chair) and sample multiple parts from the component library that could compose such a shape. 
During this process, we also generate the corresponding text caption using a text template that includes conjunctions to synthesize part-level captions into a comprehensive shape-level caption.
Given that the shape and caption generation process is parameter-free and computation-efficient, it is integrated into the training process to dynamically create new shapes and captions with diverse geometry and semantics. By doing so, we continuously obtain extensive augmented training data that contributes to model training.
It is important to note that different parts may not align well on axes, meaning the distances between sampled parts may be too far or too close. Directly assembling these parts in the same 3D space can result in shapes with poor geometry. Therefore, we adjust the inter and intra distances of parts to ensure the high quality of generated shapes. The comprehensive methodology of 3D-caption paired data generation is delineated in Algorithm \ref{alg1}.

For the inter distances among parts, we reposition the centroids to the origin and compute distances between parts along three axes. We then adjust the coordinates of parts to ensure they do not overlap and exhibit standard shape geometry (e.g., the table base should be below the top with proper margin). 
Regarding the intra distances within a part, we first illustrate an example to explain the necessity of this process. For instance, the legs of a large table may be far apart. When combining this part with a small tabletop, the intra distances within the legs should be reduced so that the top can fully cover the legs. 
In this case, we define covering as most projected points on XY\mbox{-}plane of base is within the top. If not, we move every point in the base towards the origin on the XY\mbox{-}plane with proper distance.

\subsection{Unimodal Encoders}

\paragraph{The 3D Shape Encoder} To extract 3D shape features of a point cloud shape $S$, we utilize PointNet \cite{qi2017pointnet} as the backbone. The encoded point-level features are represented as $\left\{ \hat{s}_{n} \right\}_{n = 1}^{N_p} \in \mathbb{R}^{N_p \times D}$, where $N_p$ is the total number of points and $D$ is the feature dimension. 
Similar to the segmentation head of PointNet, we then fuse the local and global information to enhance features with local geometry and global semantics. Specifically, we first obtain the aggregated shape-level feature $s^g \in \mathbb{R}^{D}$ through max-pooling.
Then each feature in $\left\{ \hat{s}_{n} \right\}_{n = 1}^{N_p}$ is concatenated with $s^g$ as $\left\{ \overline{s}_{n} \right\}_{n = 1}^{N_p}$, where $\overline{s}_{n} = [\hat{s}_{n};s^g]$.
The fused features are fed into a multilayered perceptron (MLP) with ReLU activation.
Following \cite{tang2021parts2words}, we further add a segmentation head and aggregate point features into features of segmented parts through average pooling.
The final output of the shape encoder is $\left\{ {s}_{n} \right\}_{n = 1}^{N} \in \mathbb{R}^{N \times D}$, where $N$ is the number of segmented parts.

\paragraph{The Text Encoder} We first initialize word embeddings $E = \left\{ e_{m} \right\}_{m = 1}^{M}$ of the text caption $T$, where $M$ is the word number in $T$.
Then we encode $E$ through a bi-directional Gate Recurrent Unit (GRU) \cite{chung2014empirical} to fuse the sequential information among the embeddings.
The encoded representation is denoted as $W = \left\{ w_{m} \right\}_{m = 1}^{M} \in \mathbb{R}^{M \times D}$:
\begin{equation}
\begin{split}
    h_{i}^{f}&={GRU}^{f}\left(e_{i}, h_{(i-1)}^{f}\right), \\
h_{i}^{b}&={GRU}^{b}\left(e_{i}, h_{(i+1)}^{b}\right), \\
w_{i}&=\left[h_{i}^{f} ; h_{i}^{b}\right],
\end{split}
\end{equation}
where ${GRU}^{f}$ and ${GRU}^{b}$ are the forward and backward GRU, $h_{i}^{f}$ and $h_{i}^{b}$ are the forward and backward hidden state of GRU for the $i$-th word, respectively. 
In the forward GRU, the $i$-th hidden state is computed with the $i$-th word embedding and the hidden state from the previous timestamp. Conversely, in the backward GRU, the $i$-th hidden state is calculated with the $i$-th word embedding and the hidden state from the next timestamp.

\subsection{The Matching Module}
After obtaining the unimodal features of 3D shapes and text captions, we compute the EMD scores of each pair as the cross-modal similarity. 
EMD is defined as an optimal transport problem between shapes and text captions. 
The transport cost $c_{ij}$ between EMD nodes $s_i$ and $w_j$ is defined as $1 - cos\left(s_i, w_j\right)$, where $cos$ is the cosine similarity:

\begin{equation}
    cos\left(s_i, w_j\right)=\frac{s_i^{\top} w_j}{\left\|s_i\right\| \cdot\left\|w_j\right\|}.
\end{equation}

The Sinkhorn algorithm is introduced to compute the EMD matching flow $x_{ij}$. Then the similarity score between the 3D shape $S$ and text caption $T$ is calculated as:
\begin{equation}
\label{equation::EMD_SIM}
    EMD\left(S, T\right)=-\sum_{i=1}^{N} \sum_{j=1}^{M} c_{i j} x_{i j}.
\end{equation}

By adopting EMD, we compute the fine-grained similarity between each part of 3D shapes and text captions, which models the cross-modal alignment at the local semantic level.

\subsection{The Training Objective}
We optimize the segmentation module in the shape encoder by a cross-entropy loss $L_{SEG}$.
To achieve bidirectional cross-modal retrieval and obtain discriminative features, we employ contrastive learning as the training objective. Concretely, we utilize the InfoNCE loss \cite{oord2018representation} to jointly optimize the shape-to-text (S2T) and text-to-shape (T2S) retrieval tasks. Within a batch with size of $B$, the similarity between shapes and texts of the $B$ positive pairs are maximized while minimizing the similarity of the $B^2 - B$ negative pairs:

\begin{equation}
L_{S2T} =-\frac{1}{B} \sum_{i}^{B} \log \frac{\exp \left(EMD\left( S_i, T_i\right) / \tau\right)}{\sum_{j=1}^{B} \exp \left(EMD\left( S_i, T_j\right) / \tau\right)},
\end{equation}

\begin{equation}
L_{T2S} =-\frac{1}{B} \sum_{i}^{B} \log \frac{\exp \left(EMD\left( T_i, S_i\right) / \tau\right)}{\sum_{j=1}^{B} \exp \left(EMD\left( T_i, S_j\right) / \tau\right)},
\end{equation}
where $S_i$ and $T_i$ are the $i$-th shape and text caption in a batch, $EMD$ is the similarity function defined in Equation (\ref{equation::EMD_SIM}) and $\tau$ is the temperature parameter. 
By optimizing InfoNCE, the unimodal encoders maximize the mutual information between the positive pair $(S_i, T_i)$.

The overall training objective is the sum of the above three losses:
\begin{equation}
L = L_{SEG} + L_{S2T} + L_{T2S}.
\end{equation}

\begin{table*}[t]     
\centering
  \caption{\textbf{Comparison results on the Text2Shape dataset.} S2T and T2S indicate shape-to-text and text-to-shape retrieval, respectively. We achieve state-of-the-art results across all metrics.}
  \label{table::compare_with_SOTA}
\begin{tabular}{lccccccc}
\toprule
\multicolumn{1}{l}{Method} & Venue & \multicolumn{3}{c}{S2T} & \multicolumn{3}{c}{T2S} \\ 
\cmidrule(r){3-5} \cmidrule(r){6-8} 
                            &  & RR@1   &  RR@5  & NDCG@5       & RR@1   &  RR@5  & NDCG@5      \\ \midrule
 Text2Shape \cite{chen2019text2shape}     & ACCV'2018 &  0.83        &  3.37   & 0.73   &  0.40   &  2.37 & 1.35        \\
 $\rm Y^{2}$Seq2Seq \cite{han2019y2seq2seq}     & AAAI'2019 &  6.77         &  19.30     & 5.30 & 2.93    & 9.23    & 6.05    \\
 TriCoLo \cite{ruan2024tricolo}   & WACV'2024 &  16.33          & 45.52     & 12.73  & 10.25   & 29.07     &  19.85    \\
 Parts2Words \cite{tang2021parts2words}  & CVPR'2023  & 19.38     &   47.17          & 15.30 & 12.72   & 32.98        & 23.13  \\
 COM3D \cite{wu2024com3d} & ICME'2024   & 20.03     &   48.32          & 15.62 & 13.12   & 33.48    &   23.89    \\
 \midrule
\rowcolor{green!15} SCA3D (Ours)  &  ICRA'2025  &  \textbf{27.22}    &   \textbf{55.56}   &    \textbf{19.04}      & \textbf{16.67}   & \textbf{38.90}   &   \textbf{28.17}      \\
\bottomrule
\end{tabular}
\vspace{-1.5em}
\end{table*}

\section{EXPERIMENTS}

\subsection{Experiment Setup}

\paragraph{Dataset} The Text2Shape \cite{chen2019text2shape} dataset is a subset of ShapeNet \cite{chang2015shapenet} and PartNet \cite{mo2019partnet} with additional text annotations. Following the split by \cite{tang2021parts2words}, the training and test sets contain 11,498 and 1,434 3D shapes, respectively. Each shape is associated with an average of 5 captions, allowing the model to align 3D shapes with varying text semantics. The semantic segmentation labels are provided by PartNet, specifically using the coarse granularity which consists of 17 segmentation classes.

\paragraph{Evaluation Metrics} To evaluate the cross-modal 3D retrieval task, we adopt the commonly used Recall Rate at $k$ (RR@k) and Normalized Discounted Cumulative Gain (NDCG) \cite{jarvelin2002cumulated} as metrics.
RR@k measures the proportion of relevant items that are successfully retrieved within the top-k results, where k is set to 1 and 5.
NDCG evaluates the quality of a ranking system by considering both the relevance and the position of the retrieved items.

\paragraph{Implementation Details} To extract point cloud features, we utilize PointNet \cite{qi2017pointnet} as the 3D shape encoder. Each point cloud is sampled to $N_p\!=\!2,500$ points for better computation efficiency and saving memory.
The text encoder is a single-layer bi-directional GRU and word embeddings are initialized from scratch.
The feature dimension $D$ is set to 1024. 
LLaVA-1.6-Vicuna-13B \cite{liu2024llavanext} is deployed as the MLLM. We render 3D shapes to 6 multi-view images at distinct camera positions.
The temperature parameter $\tau$ is set to 0.1
The model is trained for 90 epochs with a batch size of 128. 
Adam optimizer \cite{KingmaB14} is applied with an initial learning rate of 0.0004 and a linear decay schedule. Gradient clipping is set to 2.0 to prevent the gradient exploding.

\subsection{Comparison with State-of-the-Arts}
We compare our method with previous state-of-the-art (SOTA) methods, including Text2Shape \cite{chen2019text2shape}, $\rm Y^{2}$Seq2Seq \cite{han2019y2seq2seq}, TriCoLo \cite{ruan2024tricolo}, Parts2Words \cite{tang2021parts2words}, and COM3D \cite{wu2024com3d}.
The experimental results on the Text2Shape dataset \cite{chen2019text2shape} are summarized in Table \ref{table::compare_with_SOTA}.
Notably, our method significantly surpasses the previous SOTA method COM3D across all evaluation metrics by a substantial margin. The relative improvements range from 14.98\% to 35.90\%, demonstrating the superior effectiveness of our approach.

\subsection{Ablation Study} 

\begin{table}[t]     
\scriptsize
\tabcolsep=3.3pt
\centering
  \caption{\textbf{Ablation study on S2T and T2S tasks.} DataAug represents data augmentation. In Rows 2 and 3, EMD and InfoNCE are replaced by cosine similarity and semi-hard triplet loss, respectively.}
  \label{table::ablation study}
\resizebox{1\linewidth}{!}{
\begin{tabular}{clcccccc}
\toprule
\multicolumn{1}{l}{Row}  & \multicolumn{1}{l}{Setting} & \multicolumn{3}{c}{S2T} & \multicolumn{3}{c}{T2S} \\ 
\cmidrule(r){3-5} \cmidrule(r){6-8} 
                            &  &  RR@1   &  RR@5  & NDCG@5       & RR@1   &  RR@5  & NDCG@5      \\ \midrule
 1 & w/o DataAug     &   22.14    &  50.02           & 16.31 & 13.74   & 35.11        & 24.58  \\
 2 & w/o EMD      &   23.44    &     52.48        & 17.32 & 14.94   &  36.63   &  26.15     \\
 3 & w/o InfoNCE      & 24.35 & 53.67 & 18.01 & 15.08 & 37.12 & 26.45       \\
\rowcolor{green!15} \cellcolor{white}4 & SCA3D (Ours)      &    \textbf{27.22}    &   \textbf{55.56}   &    \textbf{19.04}      & \textbf{16.67}   & \textbf{38.90}   &   \textbf{28.17}      \\
\bottomrule
\end{tabular}
}
\vspace{-1.5em}
\end{table}

\begin{table}[t]     
\scriptsize
\tabcolsep=3.5pt
\centering
  \caption{\textbf{Ablation study on part distance adjustments.}}
  \label{table::ablation study dis adj}
\resizebox{1\linewidth}{!}{
\begin{tabular}{cccccccccc}
\toprule
\multicolumn{1}{c}{Row} & \multicolumn{2}{c}{Setting}   & \multicolumn{3}{c}{S2T} & \multicolumn{3}{c}{T2S} \\ 
\cmidrule(r){2-3} \cmidrule(r){4-6} \cmidrule(r){7-9} 
              &     inter         & intra &  RR@1   &  RR@5  & NDCG@5       & RR@1   &  RR@5  & NDCG@5      \\ \midrule
1 & \ding{56} &  \ding{56}     &   19.73 & 45.84 & 14.85 & 12.40 & 33.40 & 23.17     \\
2 & \ding{51} &  \ding{56}     &    25.05 & 53.39 & 18.19 & 15.26 & 36.63 & 26.30      \\
3 &  \ding{56} & \ding{51}      &   19.59 & 46.22 & 14.90 & 13.06 & 33.46 & 23.92     \\
\rowcolor{green!15} \cellcolor{white}4 & \ding{51} & \ding{51}      &    \textbf{27.22}    &   \textbf{55.56}   &    \textbf{19.04}      & \textbf{16.67}   & \textbf{38.90}   &   \textbf{28.17}      \\
\bottomrule
\end{tabular}
}
\vspace{-1.5em}
\end{table}

\paragraph{Data Augmentation} We validate the efficacy of our proposed data augmentation method. As illustrated in Table \ref{table::ablation study} Row 1, the retrieval accuracy significantly declines in the absence of data augmentation. The most pronounced performance degradation among all ablation studies in Table \ref{table::ablation study} indicates that the performance enhancements of our method are primarily attributable to data augmentation. This demonstrates that our data augmentation method effectively enriches the diversity of the training set, resulting in substantially improved performance.

\paragraph{Part Distance Adjustments} We assess the impact of part distance adjustments in Table \ref{table::ablation study dis adj} . All metrics significantly deteriorate as the quality of generated 3D shapes declines without the crucial adjusting process (Row 1). Many generated shapes exhibit amorphous geometric structures, leading to model confusion and introducing noise.
The application of inter (Row 2) and intra (Row 3) distance adjustments results in distinct outcomes. With inter adjustments, performance significantly improves as many generated 3D shapes begin to exhibit standard shape geometry. Conversely, applying only intra-adjustments without inter-adjustments leads to minimal improvements, as the generated shapes still exhibit poor geometry. Furthermore, their combined application results in even better performance (Row 4), as the high geometric quality of the generated shapes is ensured.

\paragraph{Similarity Function} To demonstrate the contribution of EMD as the similarity function, we replace it with cosine similarity, which is commonly used by 2D-text retrieval methods \cite{radford2021learning, li2022blip}. As shown in Table \ref{table::ablation study} Row 2, cosine similarity performs worse than EMD. The performance is limited because cosine similarity measures shapes and texts at the global level, neglecting essential local geometries and semantics. In contrast, EMD enables fine-grained cross-modal matching, which better aligns the embeddings of 3D shape and text modalities.

\paragraph{Loss Function} We validate the influence of InfoNCE as our contrastive learning loss function. Table \ref{table::ablation study} Row 3 summarizes the results of the semi-hard triplet loss adopted by Part2Words \cite{tang2021parts2words} and COM3D \cite{wu2024com3d}, which perform worse than the InfoNCE loss. Note that our method still achieves better performance than Part2Words and COM3D even with the semi-hard triplet loss, highlighting the superior effectiveness of our proposed method.

\subsection{Qualitative Results}

\begin{figure}[t]
\centering
  \subfloat{\includegraphics[width=1\linewidth]{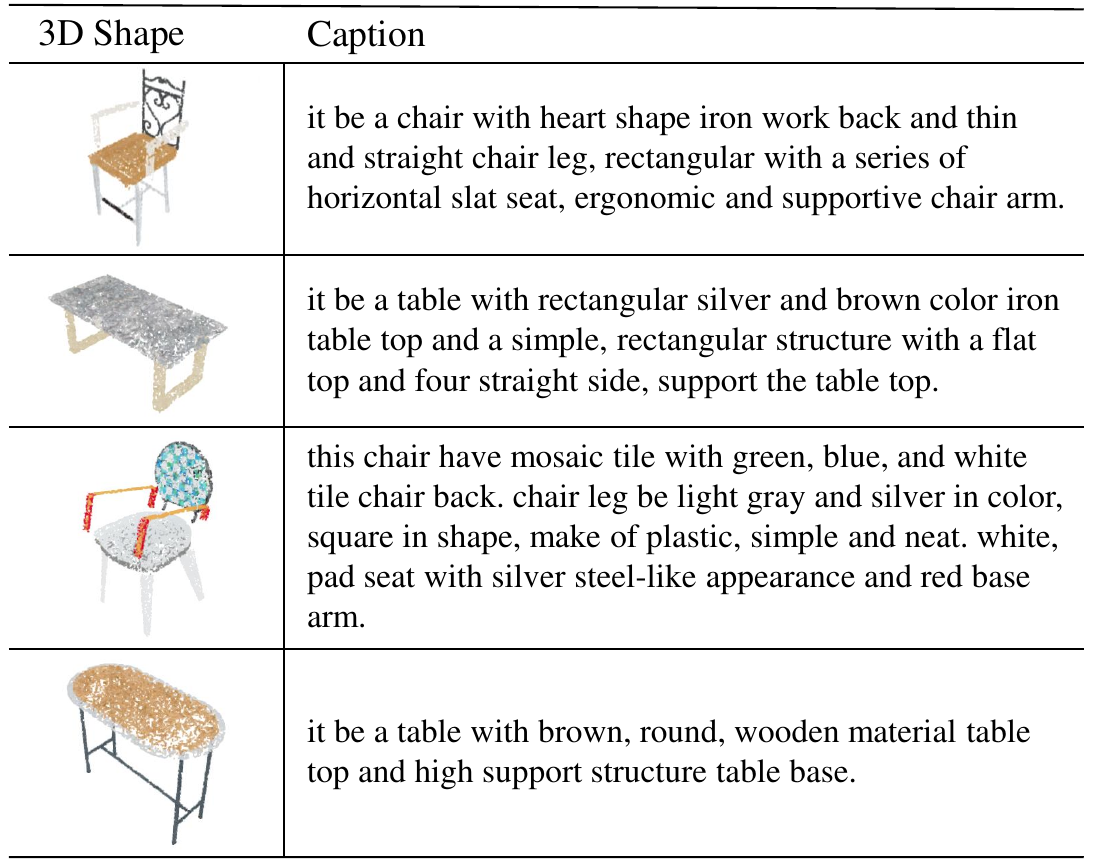}} 
   
\caption{\textbf{Generated 3D shapes and captions} through data augmentation.}
\label{figure::DataAug}
\vspace{-0.5em}
\end{figure}

\begin{figure}[t]
\centering
  \subfloat{\includegraphics[width=1\linewidth]{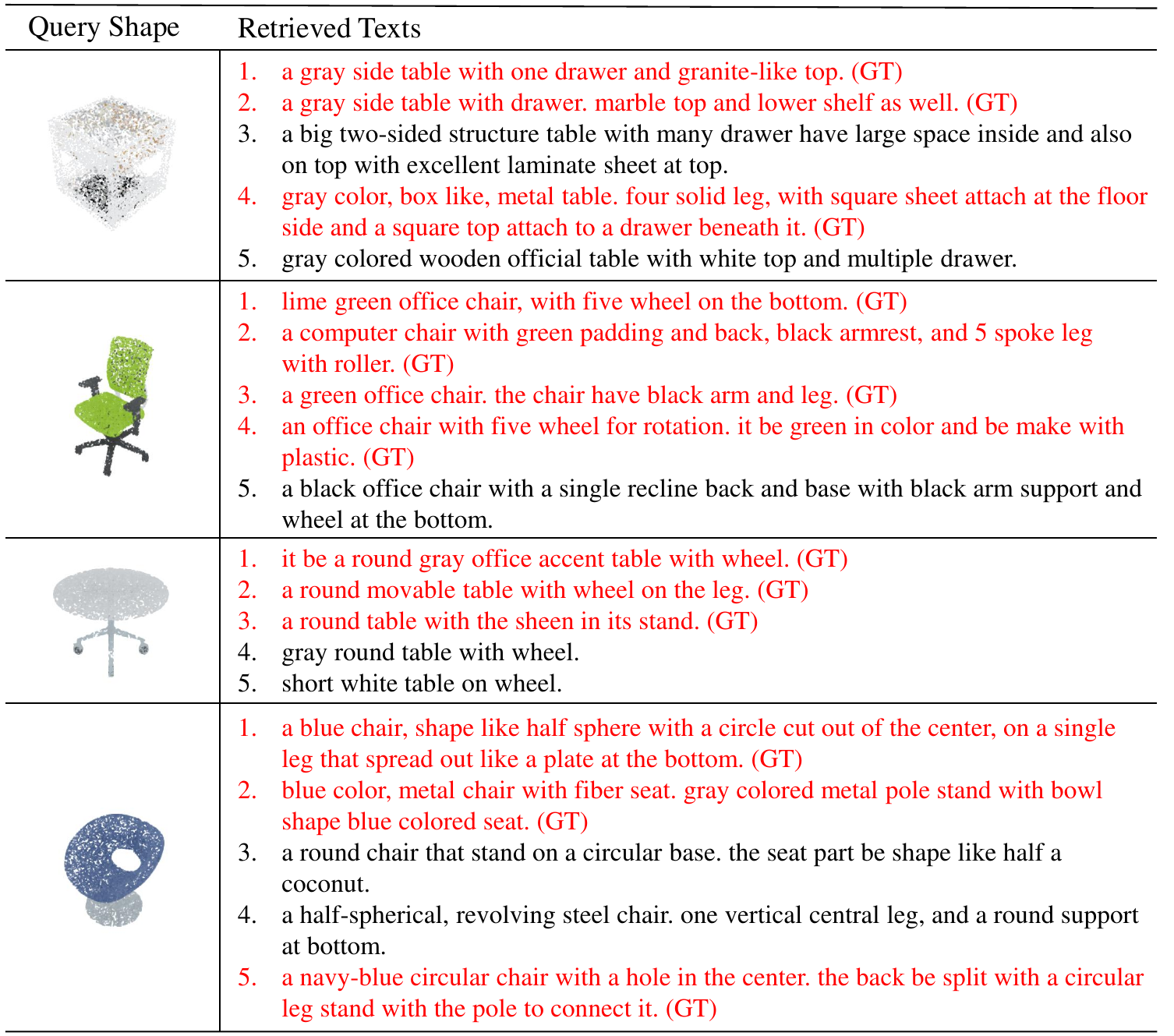}} 
   
\caption{\textbf{Shape-to-text retrieval results.} Each query shape is displayed with the top-5 ranked texts. Ground truths are highlighted in \textcolor[rgb]{1,0,0}{red}. }
\label{figure::Retrieval Results S2T}
\vspace{-1.5em}
\end{figure}

\begin{figure}[t]
\centering
  \subfloat{\includegraphics[width=1\linewidth]{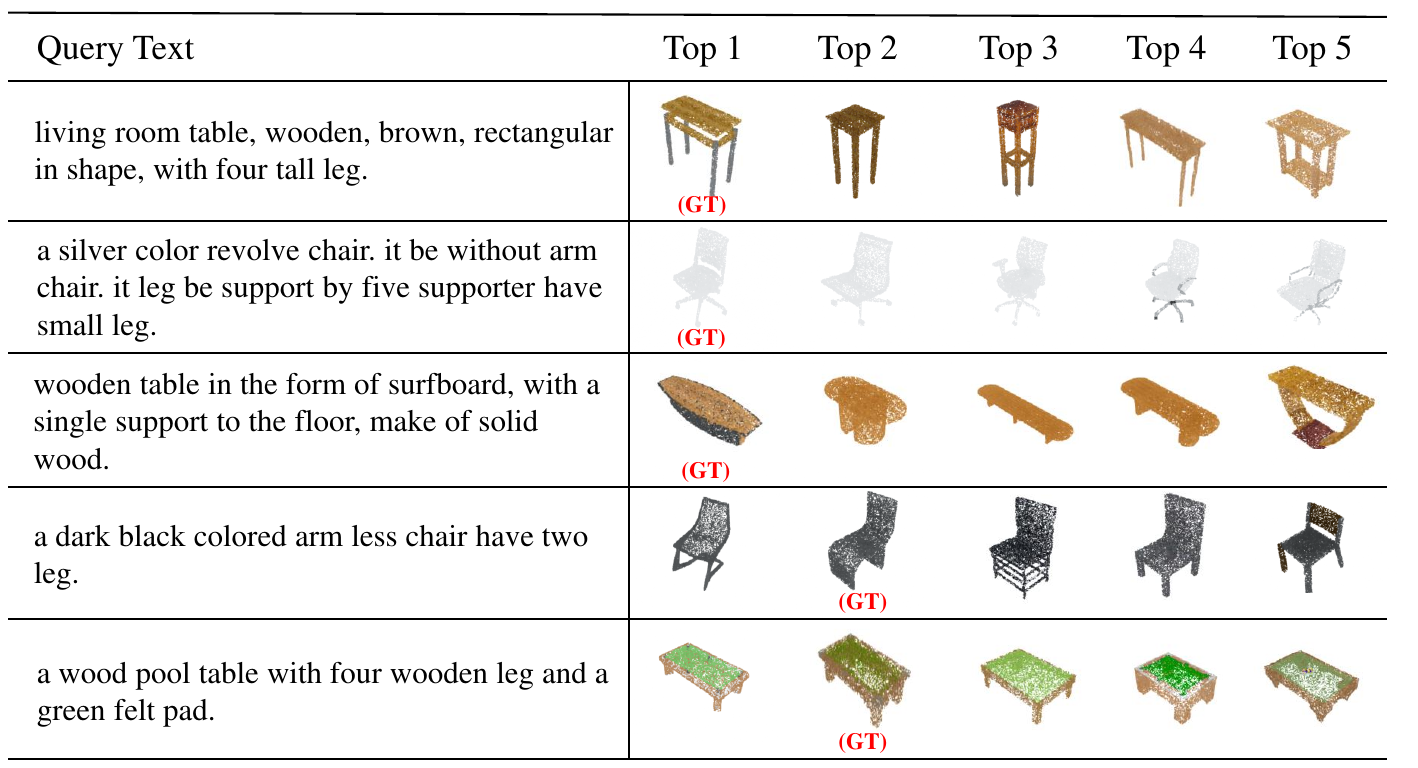}} 
   
\caption{\textbf{Text-to-shape retrieval results.} Each query text is displayed with the top-5 ranked shapes. Ground truths are indicated as \textcolor[rgb]{1,0,0}{GT}.}
\label{figure::Retrieval Results T2S}
\vspace{-1.5em}
\end{figure}

\paragraph{Generated 3D Shapes and Captions}

The visualized examples of generated 3D shapes with captions are illustrated in Fig. \ref{figure::DataAug}. Our data augmentation method creates high-quality 3D shapes with precise and contextually accurate text captions. This meticulous alignment between shapes and captions enhances the visual appeal and significantly contributes to the performance boost observed in our experiments. The enriched diversity and quality of the training data facilitated by our data augmentation technique ensure that the model learns more robust and discriminative features, leading to prominent retrieval accuracy and overall effectiveness.

\paragraph{Retrieval Results}

We present visualization examples of S2T and T2S retrieval results in Fig. \ref{figure::Retrieval Results S2T} and Fig. \ref{figure::Retrieval Results T2S}, respectively. Each query is displayed with the top-5 retrieved items. In Fig. \ref{figure::Retrieval Results S2T}, our model successfully matches the query shapes with an average of 3 ground truth texts (each shape has 5 ground truths). In Fig. \ref{figure::Retrieval Results T2S}, all retrieved shapes are highly ranked, validating the remarkable retrieval ability of our method. 
It is worth noting that almost all the retrieved items share similar semantic or geometric/color details, and even the non-ground truths align well with the queries. 
The high degree of semantic and geometric consistency among the retrieved items underscores the efficacy of our approach in capturing and leveraging the intricate relationships between 3D shapes and their textual descriptions.

\section{CONCLUSION}
We introduce a novel online data augmentation method to enhance cross-modal 3D retrieval by generating paired data of 3D shapes and textual captions. Leveraging the powerful inference capabilities of the multimodal large language model, we comprehend the geometry and semantics of each component within 3D shapes. From this understanding, we generate a vast array of new 3D shapes and their corresponding descriptions. Throughout this generation process, we optimize alignment both within each component and between components to create realistic and coherent objects. Finally, for cross-modal matching, we employ EMD similarity and contrastive learning to refine the retrieval outcomes. Extensive experiments demonstrated that our
SCA3D achieves state-of-the-art performance in both shape-to-text and text-to-shape retrieval tasks. In the future, we aim to expand our data augmentation approach across more complex 3D environments to enhance its practical application effectiveness.

\section*{ACKNOWLEDGMENT}
This research is supported by the National Natural Science Foundation of China (No. 62406267), Guangzhou-HKUST(GZ) Joint Funding Program (Grant No.2025A03J3956), Education Bureau of Guangzhou Municipality and the Guangzhou Municipal Education Project (No. 2024312122).







\bibliographystyle{IEEEtran}
\bibliography{IEEEabrv,refs}

\begin{thebibliography}{10}
\providecommand{\url}[1]{#1}
\csname url@samestyle\endcsname
\providecommand{\newblock}{\relax}
\providecommand{\bibinfo}[2]{#2}
\providecommand{\BIBentrySTDinterwordspacing}{\spaceskip=0pt\relax}
\providecommand{\BIBentryALTinterwordstretchfactor}{4}
\providecommand{\BIBentryALTinterwordspacing}{\spaceskip=\fontdimen2\font plus
\BIBentryALTinterwordstretchfactor\fontdimen3\font minus \fontdimen4\font\relax}
\providecommand{\BIBforeignlanguage}[2]{{%
\expandafter\ifx\csname l@#1\endcsname\relax
\typeout{** WARNING: IEEEtran.bst: No hyphenation pattern has been}%
\typeout{** loaded for the language `#1'. Using the pattern for}%
\typeout{** the default language instead.}%
\else
\language=\csname l@#1\endcsname
\fi
#2}}
\providecommand{\BIBdecl}{\relax}
\BIBdecl

\bibitem{huang2023visual}
C.~Huang, O.~Mees, A.~Zeng, and W.~Burgard, ``Visual language maps for robot navigation,'' in \emph{2023 IEEE International Conference on Robotics and Automation (ICRA)}.\hskip 1em plus 0.5em minus 0.4em\relax IEEE, 2023, pp. 10\,608--10\,615.

\bibitem{li2024manipllm}
X.~Li, M.~Zhang, Y.~Geng, H.~Geng, Y.~Long, Y.~Shen, R.~Zhang, J.~Liu, and H.~Dong, ``Manipllm: Embodied multimodal large language model for object-centric robotic manipulation,'' in \emph{Proceedings of the IEEE/CVF Conference on Computer Vision and Pattern Recognition}, 2024, pp. 18\,061--18\,070.

\bibitem{huang2024copa}
H.~Huang, F.~Lin, Y.~Hu, S.~Wang, and Y.~Gao, ``Copa: General robotic manipulation through spatial constraints of parts with foundation models,'' in \emph{First Workshop on Vision-Language Models for Navigation and Manipulation at ICRA 2024}, 2024.

\bibitem{hu2024look}
Y.~Hu, F.~Lin, T.~Zhang, L.~Yi, and Y.~Gao, ``Look before you leap: Unveiling the power of gpt-4v in robotic vision-language planning,'' in \emph{First Workshop on Vision-Language Models for Navigation and Manipulation at ICRA 2024}, 2024.

\bibitem{zhang2024multigo}
G.~Zhang, N.~Yao, S.~Zhang, H.~Zhao, G.~Pang, J.~Shu, and H.~Wang, ``Multigo: Towards multi-level geometry learning for monocular 3d textured human reconstruction,'' \emph{arXiv preprint arXiv:2412.03103}, 2024.

\bibitem{wang2024embodiedscan}
T.~Wang, X.~Mao, C.~Zhu, R.~Xu, R.~Lyu, P.~Li, X.~Chen, W.~Zhang, K.~Chen, T.~Xue \emph{et~al.}, ``Embodiedscan: A holistic multi-modal 3d perception suite towards embodied ai,'' in \emph{Proceedings of the IEEE/CVF Conference on Computer Vision and Pattern Recognition}, 2024, pp. 19\,757--19\,767.

\bibitem{bakrcot3dref}
E.~M. BAKR, M.~A. Mohamed, M.~Ahmed, H.~Slim, and M.~Elhoseiny, ``Cot3dref: Chain-of-thoughts data-efficient 3d visual grounding,'' in \emph{The Twelfth International Conference on Learning Representations}, 2024.

\bibitem{man2024lexicon3d}
Y.~Man, S.~Zheng, Z.~Bao, M.~Hebert, L.~Gui, and Y.-X. Wang, ``Lexicon3d: Probing visual foundation models for complex 3d scene understanding,'' in \emph{The Thirty-eighth Annual Conference on Neural Information Processing Systems}, 2024.

\bibitem{song2025gvkf}
G.~Song, C.~Cheng, and H.~Wang, ``Gvkf: Gaussian voxel kernel functions for highly efficient surface reconstruction in open scenes,'' \emph{Advances in Neural Information Processing Systems}, vol.~37, pp. 104\,792--104\,815, 2025.

\bibitem{cheng2025graphguided}
C.~Cheng, G.~Song, Y.~Yao, G.~Zhang, Q.~Zhou, and H.~Wang, ``Graph-guided scene reconstruction from images with 3d gaussian splatting,'' in \emph{The Thirteenth International Conference on Learning Representations}, 2025.

\bibitem{yu2025rgb}
S.~Yu, C.~Cheng, Y.~Zhou, X.~Yang, and H.~Wang, ``Rgb-only gaussian splatting slam for unbounded outdoor scenes,'' \emph{arXiv preprint arXiv:2502.15633}, 2025.

\bibitem{chen2019text2shape}
K.~Chen, C.~B. Choy, M.~Savva, A.~X. Chang, T.~Funkhouser, and S.~Savarese, ``Text2shape: Generating shapes from natural language by learning joint embeddings,'' in \emph{Computer Vision--ACCV 2018: 14th Asian Conference on Computer Vision, Perth, Australia, December 2--6, 2018, Revised Selected Papers, Part III 14}.\hskip 1em plus 0.5em minus 0.4em\relax Springer, 2019, pp. 100--116.

\bibitem{han2019y2seq2seq}
Z.~Han, M.~Shang, X.~Wang, Y.-S. Liu, and M.~Zwicker, ``Y2seq2seq: Cross-modal representation learning for 3d shape and text by joint reconstruction and prediction of view and word sequences,'' in \emph{Proceedings of the AAAI Conference on Artificial Intelligence}, vol.~33, no.~01, 2019, pp. 126--133.

\bibitem{ruan2024tricolo}
Y.~Ruan, H.-H. Lee, Y.~Zhang, K.~Zhang, and A.~X. Chang, ``Tricolo: Trimodal contrastive loss for text to shape retrieval,'' in \emph{Proceedings of the IEEE/CVF Winter Conference on Applications of Computer Vision}, 2024, pp. 5815--5825.

\bibitem{tang2021parts2words}
C.~Tang, X.~Yang, B.~Wu, Z.~Han, and Y.~Chang, ``Parts2words: Learning joint embedding of point clouds and texts by bidirectional matching between parts and words,'' in \emph{2023 IEEE/CVF Conference on Computer Vision and Pattern Recognition (CVPR)}, 2023, pp. 6884--6893.

\bibitem{wu2024com3d}
H.~Wu, R.~Li, H.~Wang, and H.~Xiong, ``Com3d: Leveraging cross-view correspondence and cross-modal mining for 3d retrieval,'' in \emph{2024 IEEE International Conference on Multimedia and Expo (ICME)}, 2024, pp. 1--6.

\bibitem{liu2024visual}
H.~Liu, C.~Li, Q.~Wu, and Y.~J. Lee, ``Visual instruction tuning,'' \emph{Advances in neural information processing systems}, vol.~36, 2024.

\bibitem{oord2018representation}
A.~v.~d. Oord, Y.~Li, and O.~Vinyals, ``Representation learning with contrastive predictive coding,'' \emph{arXiv preprint arXiv:1807.03748}, 2018.

\bibitem{radford2021learning}
A.~Radford, J.~W. Kim, C.~Hallacy, A.~Ramesh, G.~Goh, S.~Agarwal, G.~Sastry, A.~Askell, P.~Mishkin, J.~Clark \emph{et~al.}, ``Learning transferable visual models from natural language supervision,'' in \emph{International conference on machine learning}.\hskip 1em plus 0.5em minus 0.4em\relax PMLR, 2021, pp. 8748--8763.

\bibitem{li2022blip}
J.~Li, D.~Li, C.~Xiong, and S.~Hoi, ``Blip: Bootstrapping language-image pre-training for unified vision-language understanding and generation,'' in \emph{International conference on machine learning}.\hskip 1em plus 0.5em minus 0.4em\relax PMLR, 2022, pp. 12\,888--12\,900.

\bibitem{weng2023open}
Z.~Weng, X.~Yang, A.~Li, Z.~Wu, and Y.-G. Jiang, ``Open-vclip: Transforming clip to an open-vocabulary video model via interpolated weight optimization,'' in \emph{International Conference on Machine Learning}.\hskip 1em plus 0.5em minus 0.4em\relax PMLR, 2023, pp. 36\,978--36\,989.

\bibitem{schuhmann2021laion}
C.~Schuhmann, R.~Kaczmarczyk, A.~Komatsuzaki, A.~Katta, R.~Vencu, R.~Beaumont, J.~Jitsev, T.~Coombes, and C.~Mullis, ``Laion-400m: Open dataset of clip-filtered 400 million image-text pairs,'' in \emph{NeurIPS Workshop Datacentric AI}, no. FZJ-2022-00923.\hskip 1em plus 0.5em minus 0.4em\relax J{\"u}lich Supercomputing Center, 2021.

\bibitem{miech2019howto100m}
A.~Miech, D.~Zhukov, J.-B. Alayrac, M.~Tapaswi, I.~Laptev, and J.~Sivic, ``Howto100m: Learning a text-video embedding by watching hundred million narrated video clips,'' in \emph{Proceedings of the IEEE/CVF international conference on computer vision}, 2019, pp. 2630--2640.

\bibitem{deng2009imagenet}
J.~Deng, W.~Dong, R.~Socher, L.-J. Li, K.~Li, and L.~Fei-Fei, ``Imagenet: A large-scale hierarchical image database,'' in \emph{2009 IEEE conference on computer vision and pattern recognition}.\hskip 1em plus 0.5em minus 0.4em\relax Ieee, 2009, pp. 248--255.

\bibitem{Peng_2024_CVPR}
W.~Peng, S.~Xie, Z.~You, S.~Lan, and Z.~Wu, ``Synthesize diagnose and optimize: Towards fine-grained vision-language understanding,'' in \emph{Proceedings of the IEEE/CVF Conference on Computer Vision and Pattern Recognition (CVPR)}, June 2024, pp. 13\,279--13\,288.

\bibitem{doveh2023teaching}
S.~Doveh, A.~Arbelle, S.~Harary, E.~Schwartz, R.~Herzig, R.~Giryes, R.~Feris, R.~Panda, S.~Ullman, and L.~Karlinsky, ``Teaching structured vision \& language concepts to vision \& language models,'' in \emph{Proceedings of the IEEE/CVF Conference on Computer Vision and Pattern Recognition}, 2023, pp. 2657--2668.

\bibitem{podellsdxl}
D.~Podell, Z.~English, K.~Lacey, A.~Blattmann, T.~Dockhorn, J.~M{\"u}ller, J.~Penna, and R.~Rombach, ``Sdxl: Improving latent diffusion models for high-resolution image synthesis,'' in \emph{The Twelfth International Conference on Learning Representations}, 2024.

\bibitem{le2022bloom}
T.~Le~Scao, A.~Fan, C.~Akiki, E.~Pavlick, S.~Ili{\'c}, D.~Hesslow, R.~Castagn{\'e}, A.~Sasha~Luccioni, F.~Yvon, M.~Gall{\'e} \emph{et~al.}, ``Bloom: A 176b-parameter open-access multilingual language model,'' \emph{arXiv e-prints}, pp. arXiv--2211, 2022.

\bibitem{chang2015shapenet}
A.~X. Chang, T.~Funkhouser, L.~Guibas, P.~Hanrahan, Q.~Huang, Z.~Li, S.~Savarese, M.~Savva, S.~Song, H.~Su \emph{et~al.}, ``Shapenet: An information-rich 3d model repository,'' \emph{arXiv preprint arXiv:1512.03012}, 2015.

\bibitem{chung2014empirical}
J.~Chung, C.~Gulcehre, K.~Cho, and Y.~Bengio, ``Empirical evaluation of gated recurrent neural networks on sequence modeling,'' in \emph{NIPS 2014 Workshop on Deep Learning, December 2014}, 2014.

\bibitem{sajjadi2022scene}
M.~S. Sajjadi, H.~Meyer, E.~Pot, U.~Bergmann, K.~Greff, N.~Radwan, S.~Vora, M.~Lu{\v{c}}i{\'c}, D.~Duckworth, A.~Dosovitskiy \emph{et~al.}, ``Scene representation transformer: Geometry-free novel view synthesis through set-latent scene representations,'' in \emph{Proceedings of the IEEE/CVF Conference on Computer Vision and Pattern Recognition}, 2022, pp. 6229--6238.

\bibitem{zhang2022pointclip}
R.~Zhang, Z.~Guo, W.~Zhang, K.~Li, X.~Miao, B.~Cui, Y.~Qiao, P.~Gao, and H.~Li, ``Pointclip: Point cloud understanding by clip,'' in \emph{Proceedings of the IEEE/CVF conference on computer vision and pattern recognition}, 2022, pp. 8552--8562.

\bibitem{huang2023clip2point}
T.~Huang, B.~Dong, Y.~Yang, X.~Huang, R.~W. Lau, W.~Ouyang, and W.~Zuo, ``Clip2point: Transfer clip to point cloud classification with image-depth pre-training,'' in \emph{Proceedings of the IEEE/CVF International Conference on Computer Vision}, 2023, pp. 22\,157--22\,167.

\bibitem{mo2019partnet}
K.~Mo, S.~Zhu, A.~X. Chang, L.~Yi, S.~Tripathi, L.~J. Guibas, and H.~Su, ``Partnet: A large-scale benchmark for fine-grained and hierarchical part-level 3d object understanding,'' in \emph{Proceedings of the IEEE/CVF conference on computer vision and pattern recognition}, 2019, pp. 909--918.

\bibitem{qi2017pointnet}
C.~R. Qi, H.~Su, K.~Mo, and L.~J. Guibas, ``Pointnet: Deep learning on point sets for 3d classification and segmentation,'' in \emph{Proceedings of the IEEE conference on computer vision and pattern recognition}, 2017, pp. 652--660.

\bibitem{jarvelin2002cumulated}
K.~J{\"a}rvelin and J.~Kek{\"a}l{\"a}inen, ``Cumulated gain-based evaluation of ir techniques,'' \emph{ACM Transactions on Information Systems (TOIS)}, vol.~20, no.~4, pp. 422--446, 2002.

\bibitem{liu2024llavanext}
\BIBentryALTinterwordspacing
H.~Liu, C.~Li, Y.~Li, B.~Li, Y.~Zhang, S.~Shen, and Y.~J. Lee, ``Llava-next: Improved reasoning, ocr, and world knowledge,'' January 2024. [Online]. Available: \url{https://llava-vl.github.io/blog/2024-01-30-llava-next/}
\BIBentrySTDinterwordspacing

\bibitem{KingmaB14}
D.~P. Kingma and J.~Ba, ``Adam: {A} method for stochastic optimization,'' in \emph{3rd International Conference on Learning Representations}, 2015.

\end{thebibliography}

\end{document}